\documentclass[fleqn,10pt,table]{wlscirep}

\usepackage[utf8]{inputenc}
\usepackage[T1]{fontenc}
\usepackage{wrapfig}
\usepackage{multirow}
\usepackage{xcolor}

\usepackage{graphicx}
\usepackage{microtype}
\definecolor{up}{RGB}{249,236,242}
\definecolor{left}{RGB}{235,245,239}
\definecolor{diag}{RGB}{255,255,211}

\title{WhoAmI: An Automatic Tool for Visual Recognition of  Tiger and Leopard Individuals in the Wild}

\author[1,3]{Rita Pucci}
\author[2]{Jitendra Shankaraiah}
\author[2]{Devcharan Jathanna}
\author[2]{Ullas Karanth}
\author[3]{Kartic Subr}
\affil[1]{University of Udine, Italy} 
\affil[2]{Wildlife Conservation Society, India}
\affil[3]{University of Edinburgh, United Kingdom}
\date{}                     

\begin{abstract}
Photographs of wild animals in their natural habitats can be recorded unobtrusively via cameras that are triggered by motion nearby.
The installation of such \textit {camera traps} is becoming increasingly common across the world. Although this is a convenient source of invaluable data for biologists, ecologists and conservationists, the arduous task of poring through potentially millions of pictures each season introduces prohibitive costs and frustrating delays. We develop automatic algorithms that are able to \emph{detect} animals, \emph{identify the species} of animals and to \emph{recognize individual animals} for two species. we propose the first fully-automatic tool that can recognize specific individuals of leopard and tiger due to their characteristic body markings. We adopt a class of \textit{supervised learning} approach of machine learning where a Deep Convolutional Neural Network (DCNN) is trained using several instances of manually-labelled images for each of the three classification tasks. We demonstrate the effectiveness of our approach on a data set of camera-trap images recorded in the jungles of Southern India. 

\end{abstract}
\begin{document}

\flushbottom
\maketitle
\thispagestyle{empty}

\section*{Introduction}
Pictures recorded by cameras installed in wild habitats have served as windows into the lives of wild animals for many decades. Modern  \emph{camera traps} operate by combining infrared sensors with cameras so that images are automatically captured when motion is detected nearby nearby. Although this allows them to be non-intrusive and inexpensive for large areas, camera traps often capture spurious data such as leaves blowing past the sensor. Despite this, millions of photographs produced each year are used to inform several studies~\cite{}. Currently, the perusal of these pictures by human experts~\cite{} is the most effective and popular way to extract information from the data. Manual sorting is cumbersome and scales poorly with the volume of data. The ability to automatically extract information, from millions of unorganized photographs of animals, can have a profound impact on studies in animal biology, ecology and conservation.

Advancements in Machine Learning algorithms as well as Computer Vision techniques, have resulted in a large number of algorithms that can automatically extract information from visual data. Unfortunately, images from camera traps present insurmountable combinations of difficulties including insufficient light, poor framerates, cluttered backgrounds and significant occlusion. Early solutions coped with these problems by introducing a manual step and by tailoring the solution specifically to certain characteristic visual patterns in the data. For example, \textit{Extract Compare}~\cite{} was originally proposed for the identification of individual tigers using a manual mark-up of silhouettes of an unknown individual. It used a 3D database of scanned shapes of tigers and their associated textures and images. Although the software tool is currently available for several species, its generalization (originally developed for tigers) to other species required extensive effort and research~\cite{}. 

Fully-automatic methods have been developed to detect the presence of animals in camera trap images. Specifically, the use of Deep Convolutional Neural Networks (DCNN) has been shown to be effective with large datasets~\cite{Norouzzadeh} (millions of training images) as well as smaller datasets~\cite{Chen} (hundreds). One of the early adopters of DCNNs for camera trap data, Yu et al\cite{Yu}, used a linear Support Vector Model (SVM) in conjunction with local features extracted via sparse coding spatial pyramid matching (Yang\cite{Yang}), from cropped images. Although they achieve 82\% accuracy in detecting animals, this requires some manual preprocessing (cropping). Another approach to animal detection is to compute the difference of images from the same camera trap within a short period time\cite{Figueroa}. This method is useful for identify the presence of marked animals (such as felines) that can be observed in a sequence of images. 

There has been some work on species classification using DCNN. Chen et al\cite{Chen} develop a fully automatic method for classification across 20 species from North America, using a dataset of about 20,000 images. The images were used to train a DCNN that achieves an overall species classification accuracy of $38.31\%$. It is well known that performance of DCNNs depends heavily on the volume of supervisory training data\cite{Glorot}. More recently Norouzzadeh et al\cite{Norouzzadeh} developed a method that achieves an accuracy of $96.8\%$ using the VGG model (Simonyan\cite{simonyan}).

While many of the methods discussed above are useful in detecting the presence of an animal or identifying the species of the detected animal, there is a paucity of automatic techniques that can recognise individuals. For tigers, other data such as pug-marks~\cite{} and roars~\cite{} have been used to perform recognition with $90\%$ success rate. To the best of our knowledge, this paper introduces the first fully automatic method for recognition of individuals from camera trap images.  We present a unified methodology that also performs animal detection and species identification at rates comparable to existing methods. The main contribution is that our method performs individual recognition of tigers and leopards with limited training data. 




\section{Results} \label{Results}
We present the effectiveness of our classifiers for detecting animals, identifying the species of animals and for recognition of individual tiger and leopard in different subsections. We assessed our proposed classifiers quantitatively by calculating the proportion of correct predictions. We used four statistical measures that consider different combinations of true positives and true negatives: sensitivity, specificity, precision and accuracy. \textit{Sensitivity} is the fraction of images from a particular class that are correctly classified as belonging to that class. \textit{Specificity} is the fraction of images that are correctly identified as not belonging to a particular class. \textit{Precision} is the fraction of images reported to be of a particular class that are correctly identified. \textit{Accuracy} is the fraction of correct classifications, positive or negative.  If $TP$, $TN$, $FP$ and $FN$ denote the numbers of true positives, true negatives, false positives and false negatives respectively, then the measures are calculated using the following formulae: 

\begin{center}
\begin{tabular}{c c c c}
 $sensitivity = \frac{TP}{TP + FN}$\;, & 
 $specificity = \frac{TN}{TN + FP}$\;, & 
 $precision = \frac{TP}{TP + FP}$\; and &
 $accuracy = \frac{TP + TN} {TP + TN + FP + FN}$\;. \\
\end{tabular}
\end{center}
 
\subsection{Animal detection}
\begin{figure}[htbp]
  \begin{center}
    \begin{tabular}{@{}c@{}c@{}c@{}}
        \includegraphics[width=.33\linewidth]{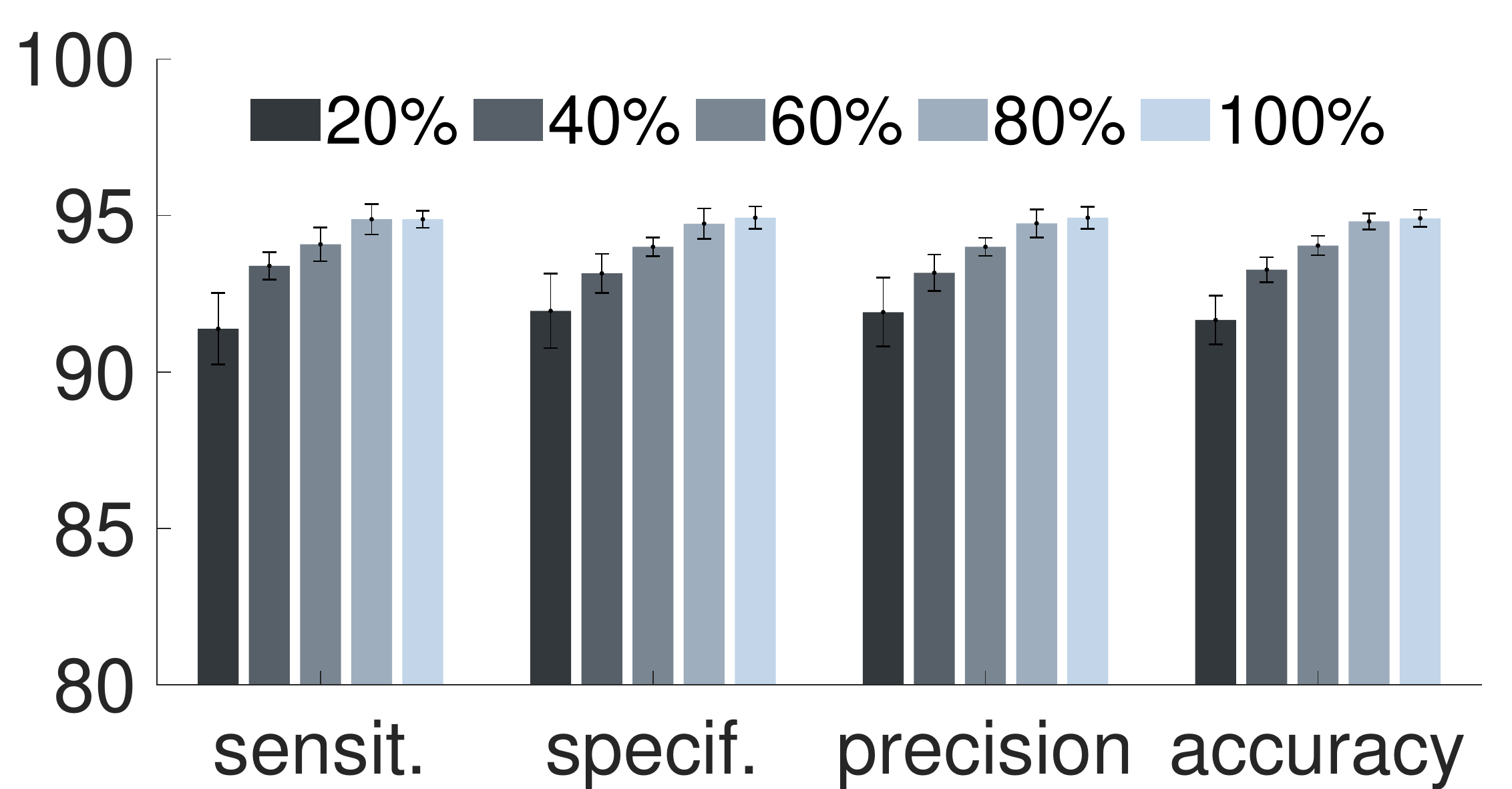} & 
        \includegraphics[width=.33\linewidth]{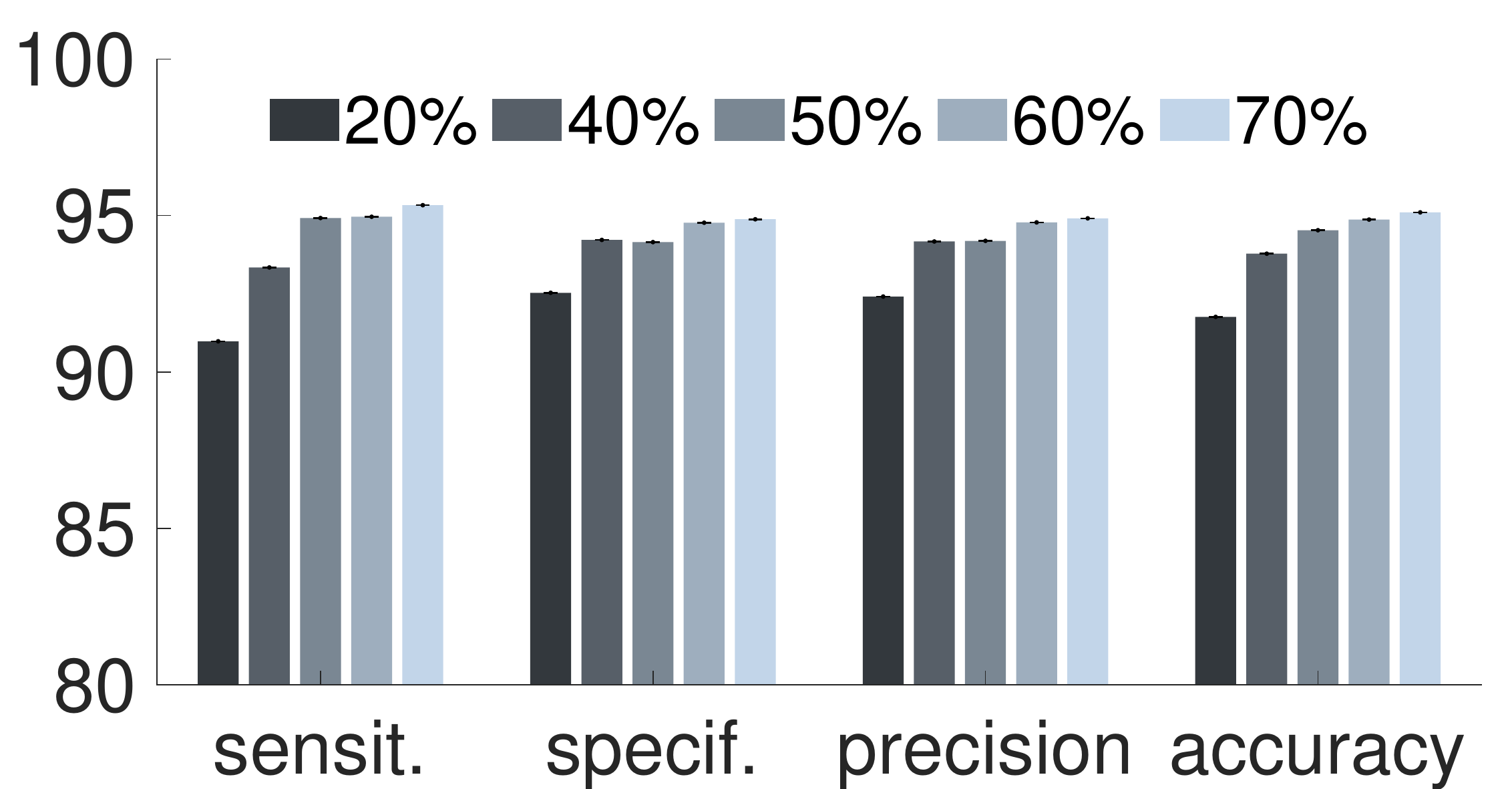} &
        \includegraphics[width=.33\linewidth]{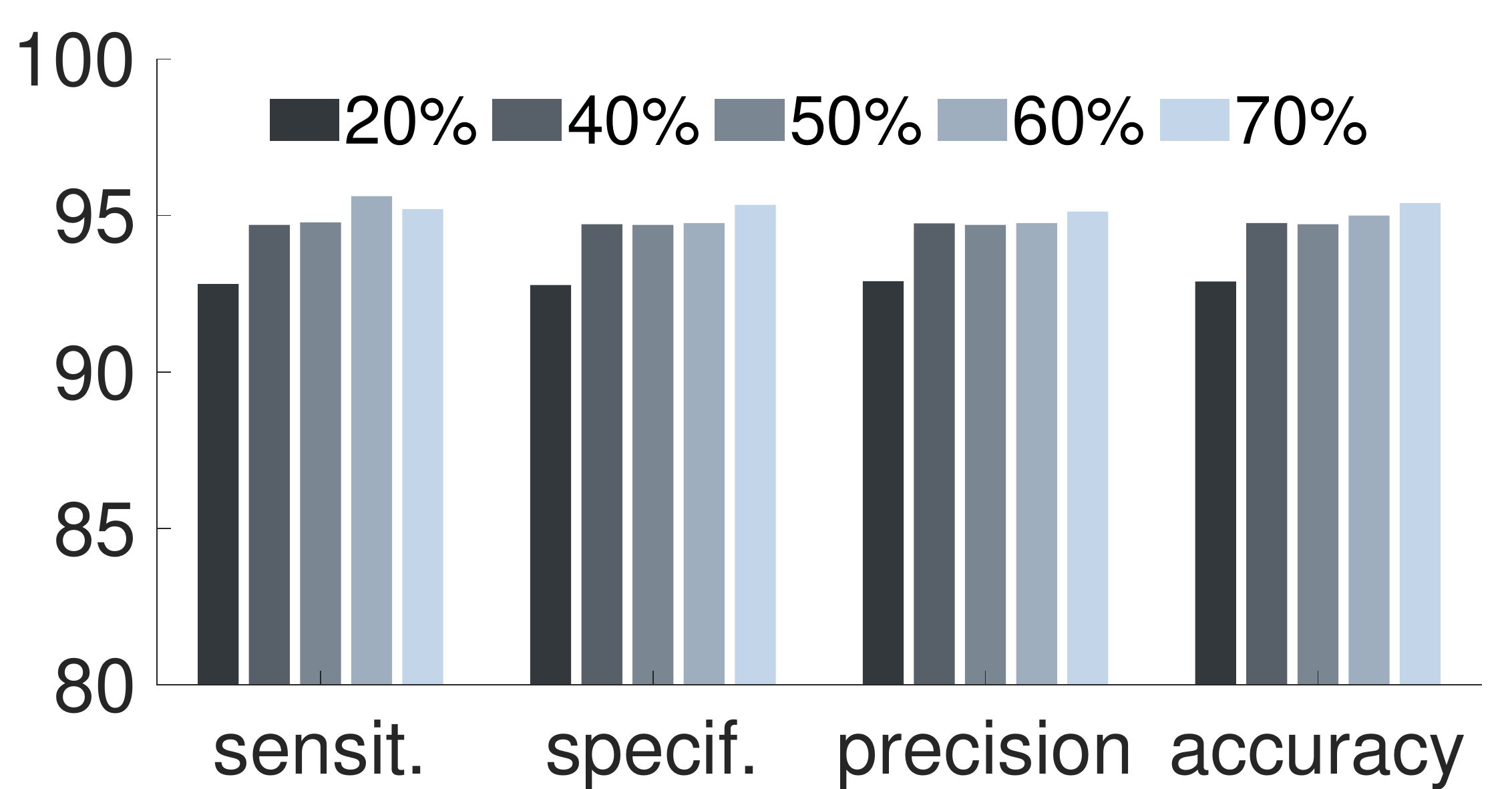} \\
        (a) volume of dataset & (b) proportion of training & (c) increasing training (fixed test)
    \end{tabular}    
    \caption{\label{fig:detection} Results for automatic animal detection as the volume of the dataset is increased (a); the percentage of the dataset used for training is varied (b); and the volume of training is increased for a fixed test set (c). }
  \end{center}
\end{figure}

Our method is able to detect animals with an accuracy of about $95\%$. Although it is possible that other state-of-the-art detectors may outperform our method on this task alone, the true strength of our approach is added abilities such as classifing species and recognising individuals.  In our experiments we observed that, for detecting the presence of animals in images, indeed increasing the volume of the dataset (Figure~\ref{fig:detection}a) leads to improvement of all measures. It is reassuring to note that with only $20\%$ of the images in the dataset, all measures are already better than $90\%$. It was also observed that training with fewer examples yields better results when tested on a large test set (Figure~\ref{fig:detection}b). Finally, we observed that a $70\%:30\%$ split of the dataset between training and test yields the best results (Figure~\ref{fig:detection}b). So we adopt this split for all other results in this paper. Details of the experimental setup used to obtain these results are explained in Section~\ref{sec:method-detection}.

\subsection{Species identification}
\begin{figure}[h]
    \begin{tabular}{cc}
        \includegraphics[width=.5\linewidth]{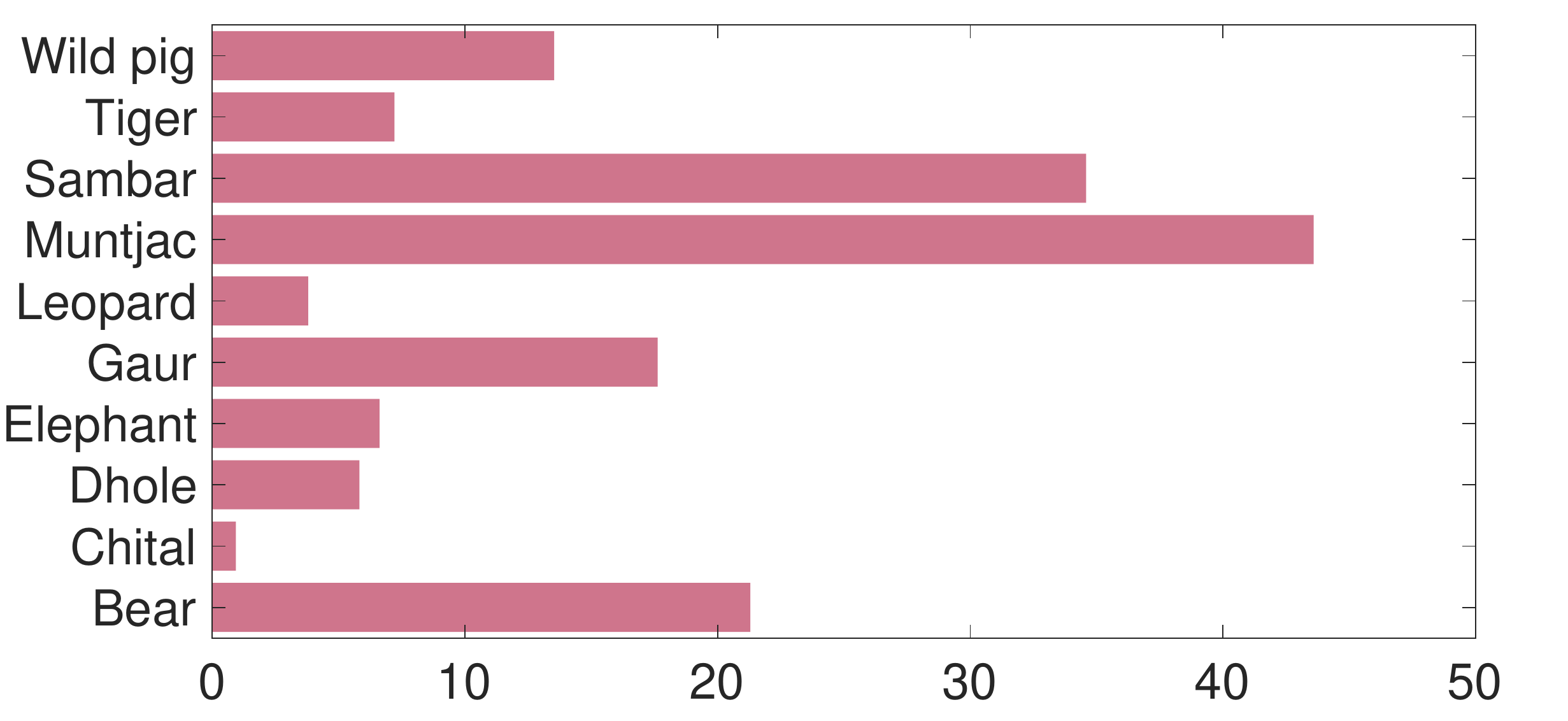}&
        \includegraphics[width=.5\linewidth]{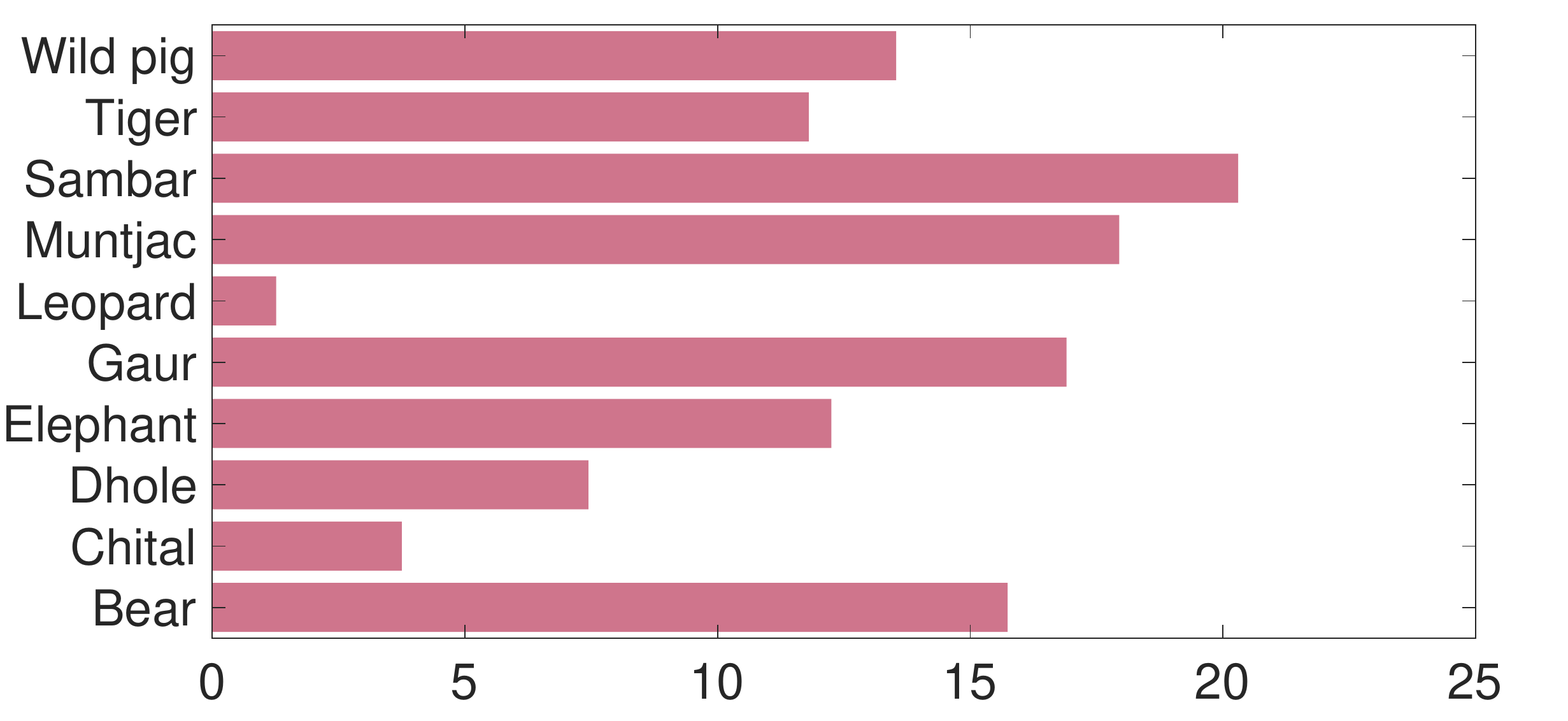} \\
        (a) False positive rate: $\frac {FP}{TP}\%$ & (b) False negative rate: $\frac{FN}{TP}\%$
    \end{tabular}
  \caption{\label{fig:SpeciesIDResult} The false positive and false negative rates for species identification show that leopards and chital (deer) are classified with least confusion.}
\end{figure}

When our method was applied to classify images based on the species of the animal present, we observed that the resulting accuracy varies widely across species. Even for species with similar numbers of training examples, the rate of false positives and false negatives vary widely. While we observed low confusion for certain species (leopard, chital, dhole, tiger), our method was less effective in distinguishing animals which lack distinguishing patterns (elephant, sambar, muntjac, bear). Figure~\ref{fig:SpeciesIDResult} shows the false positive and false negative rates across different species. The full confusion matrix is shown in Figure~\ref{fig:identification}.

\subsection{Individual recognition}
\begin{figure}[h]
    \begin{tabular}{cc}
        \includegraphics[width=.5\linewidth]{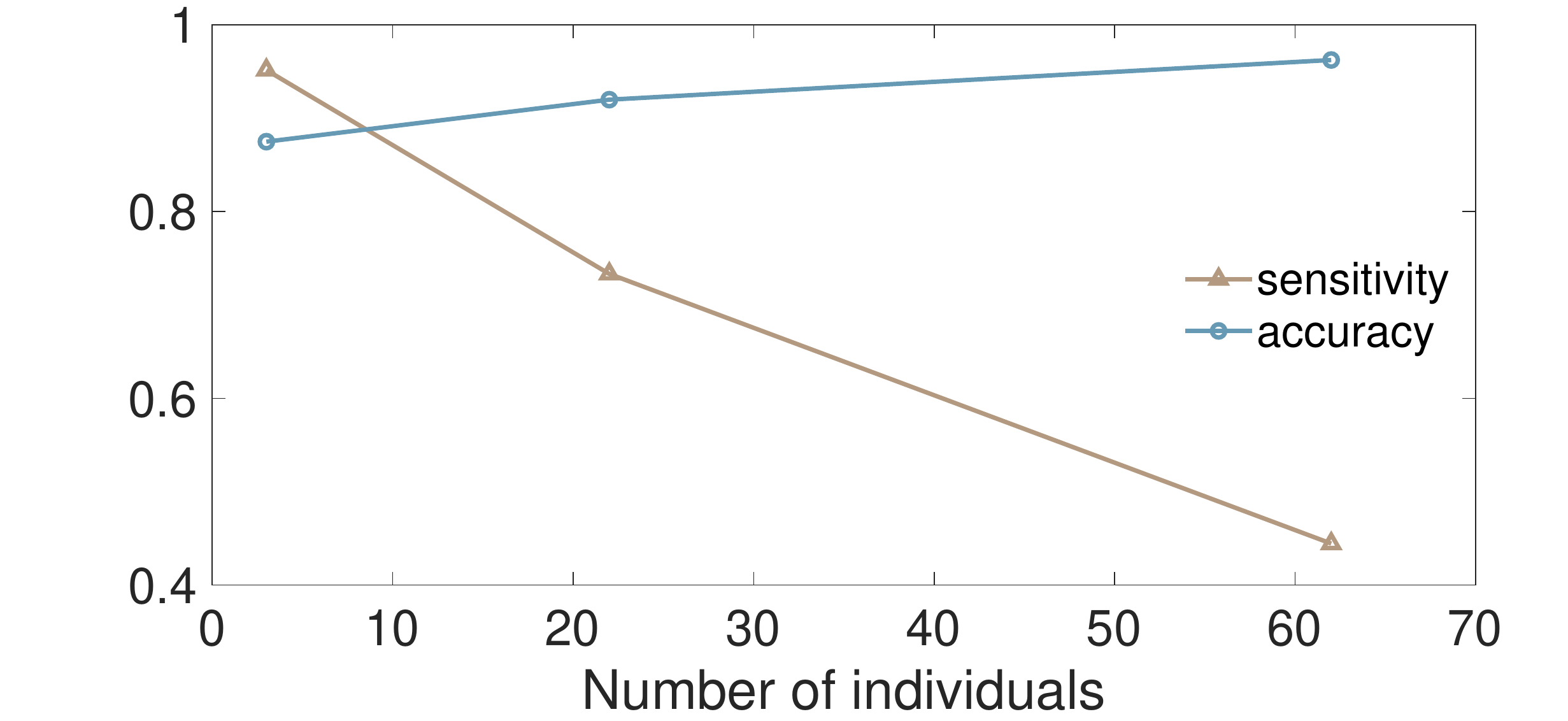}&
        \includegraphics[width=.5\linewidth]{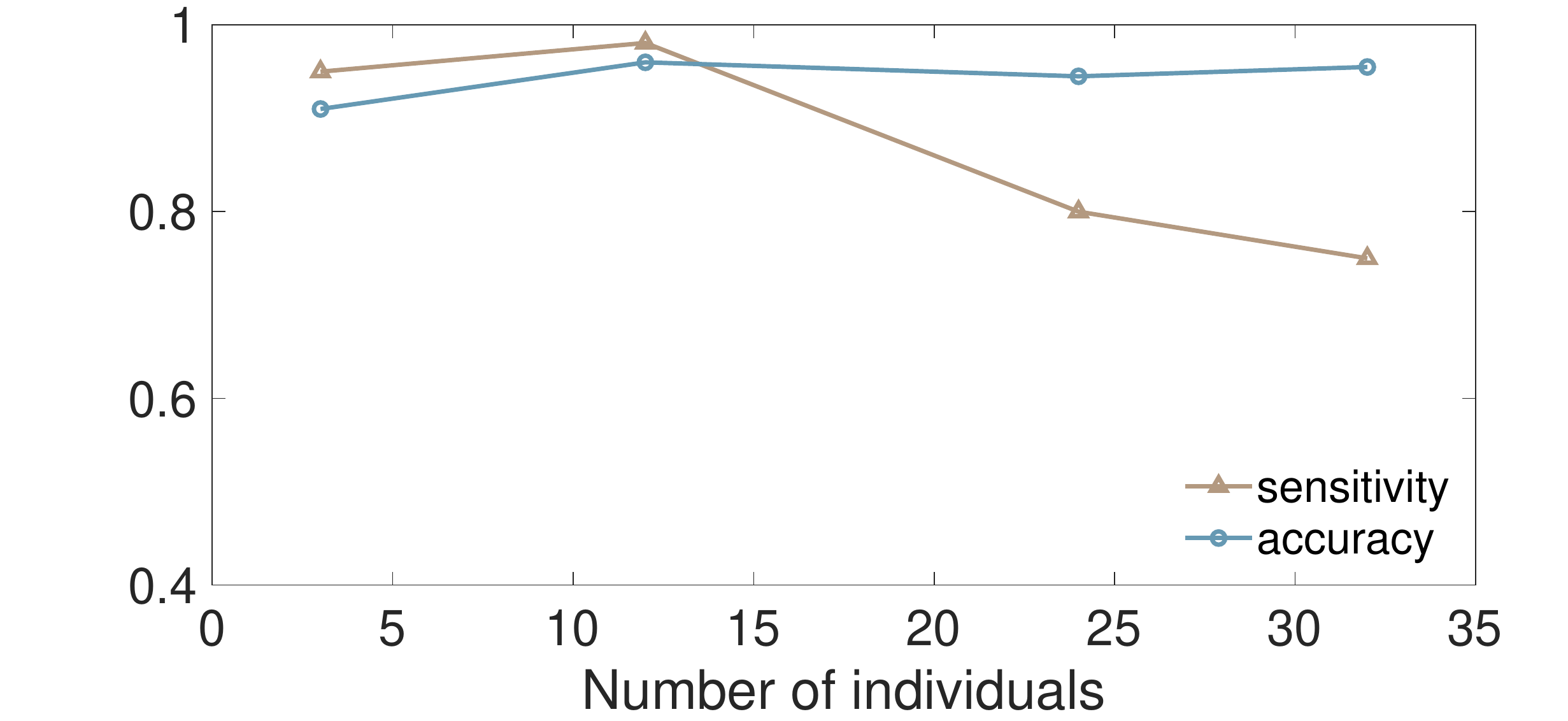} \\
        (a) leopard & (b) tiger
    \end{tabular}
  \caption{\label{fig:IndivResult} The performance of our classifier for individual leopard (a) and tiger (b) when trained with databases containing varying numbers of training classes or individuals (X-axes).}
\end{figure}

Our method is able to recognise individual animals from two species (leopard and tiger) with an accuracy of about $90\%$. We performed several tests across different training sets containing different numbers of individuals as permitted by the data (up to 62 leopards and 32 tigers). Figure~\ref{fig:IndivResult} shows plots of accuracy and sensitivity when our method was trained with a balanced dataset (similar numbers of different individuals) for leopard (a) and tiger (b).

\section*{Discussion}

\section{Methodology}
\label{Method}
We use a corpus of camera-trap images that have been tagged by human experts -- as belonging to a particular species and, where possible, an individual within the species -- to train  a machine learning algorithm to replicate the tags when presented with only the images.  In this context, learning is defined~\cite{Mohri} as progressive improvement of performance on the specific classification task.
In the rest of this section, we explain how the data needs to be  pre-processed and organised, the specific machine learning tool that we found to be most effective for the task and details on how we performed individual recognition on photographs of tigers and leopards.

\subsection{Dataset}
The dataset used in this work was collected by the Wildlife Conservation Society(WCS) India via efforts over  decades to install more than 258 camera traps across the jungles of Southern India. The  photographs used in this paper have a resolution of 2,048x1,536 pixels and those captured at night time are illuminated by a flash.  These images were  meticulously labelled by  volunteers so that each image has a tag that identifies the species of animal present in it (if any). In addition, databases of specific individuals (identified by characteristic skin markings) were created for for two species (tiger and leopard).  Experts assigned unique identification tags (names) for tiger and leopard individuals using a software tool~\cite{Hiby} that performs pattern recognition on skin markings.  Similar datasets have been used in previous studies~\cite{karanth1998,royle}.

Our dataset consists of a total of 19512 images, of which 9070 contain animals from ten species of interest (for this work):  bear (\textit{Melursus ursinus}), chital (\textit{Axis axis}), dhole (\textit{Cuon alpinus}), elephant (\textit{Elephas maximus}), gaur (\textit{Bos gaurus}), leopard (\textit{Panthera pardus fusca}), muntjac (\textit{Muntiacus}), sambar (\textit{Rusa unicolor}), tiger (\textit{Panthera tigris}), and wild pig (\textit{Sus scrofa}). Table \ref{table:disImagesSpecies} summarizes the distribution of images of the dataset among species. We labelled images that did not contain an animal from the above list as \textit{Unclassified} and considered them as negative examples. Unclassified images are those taken by camera traps when triggered by  humans who inhabit neighbouring villages, vehicles of rangers or other animals which we do not consider for this study (i.e. dogs, hares, porcupine, etc.). 
\begin{table}[ht]
	\centering
	\begin{tabular}{| l | c c c c c c c c c c|}
		\hline
		\rowcolor{up}
		&\color{black}Bear&\color{black}Chital&\color{black}Dhole&\color{black}Elephant&\color{black}Gaur&\color{black}Leopard&\color{black}Muntjac&\color{black}Sambar&\color{black}Tiger&\color{black}Wild Pig \\
		\hline
		Images & $305$ & $796$&$764$&$950$&$824$&$941$&$139$&$421$&$849$&$361$\\
		\hline
	\end{tabular}
	\caption{\label{table:disImagesSpecies} Numbers of images of each species present in our dataset.}
\end{table}

\begin{table}[ht]
	\centering
	\begin{tabular}{| r | c |  c | c | c | c |}
		\hline
		\rowcolor{up}
		\color{black}Species &
		\color{black} > $100$ images & 
		\color{black}$70$ to $100$ images &
		\color{black}$50$ to $70$ images&
		\color{black} < $50$ images & 
		\color{black} Total \\ 
		\hline
	    tiger & 3 & 9 & 12 & 8 & 32 \\
	    leopard & 21 & 21 & 21 & 0 & 63  \\
		\hline
	\end{tabular}
	\caption{\label{table:disImages} The table shows the numbers of images available in the dataset for various individuals of tiger and leopard. We had access to labelled images for 32 tiger individuals. There were no leopard individuals for which we had fewer than 70 camera-trap images.} 
\end{table}

 For animal detection, we observed better performance with a balanced dataset (9070 with animals and 9070 unclassified images). Finally, for training and evaluating our classifier that performs individual recognition, the dataset contains a total of $2317$ tiger images with varying numbers of images of each of the $32$ individuals. Three of the individuals are represented by more than $100$ images each, twelve individuals are seen in more than $70$ images each and there are at least $50$ images for $24$ individuals, this distribution is summarized in Table \ref{table:disImages}. There are $6187$ images of $62$ leopards, of which $21$ individuals are captured in more than 100 photographs and $21$ individuals are in at least $70$ images each, summarized in Table \ref{table:disImages}.

\subsection{Animal detection}
Detecting the presence of animals in images is a routine binary classification task commonly encountered in computer vision applications. We found that a widely used architecture called AlexNet~\cite{}, pre-trained on a large database of general images (ImageNet), is well suited to this task. We appended a Support Vector Machine (SVM) with a linear kernel to the output of AlexNet and trained this additional layer using our dataset. For this we found that using an equal number of positive and negative examples (9070 images with animals and 9070 without) yielded best results.

We observed that the classification accuracy was sensitive to whether or not the images were captured in daylight. To investigate this, we conducted three separate sets of training and validation experiments. In the first (diurnal), we used 3995 images of animals in daylight (and 3995 unclassified). The second set consisted of 1178 images of animals at night (and 1178 unclassified). Finally, we considered all images mixed (9070 with animals and 9070 without). For each set, we separated the images into training and validation with a split of $70\%:30\%$. The accuracies were $95.33\%$, $90.65\%$ and $90.75\%$ respectively. Although the lower accuracy for nighttime images could be attributed to using fewer images in training, it is unlikely to be the only cause since using a smaller set of diurnal images  outperforms the larger set with nighttime images mixed in. These results are presented in Table~\ref{table:nocturnal}.

\begin{table}[ht]
	\centering
	\begin{tabular}{| r | c  c c |}
		\hline
		\rowcolor{up}
		\color{black}Sub-dataset & 
		\color{black}Num. Images & 
		\color{black}Training Acc & 
		\color{black}Test Acc\\
		\hline
		\cellcolor{left}\color{black}daylight & 
		3995&
		$99.93\%$ & 
		$95.33\%$\\
		\cellcolor{left}\color{black}night & 
		1178 &
		$99.03\%$ & 
		$90.65\%$\\
		\cellcolor{left}\color{black}mixed & 
		9070 &
		$95.35\%$ & 
		$93.27\%$\\
		\hline
	\end{tabular}
	\caption{\label{table:nocturnal}Test results for images captured at different times of the day.}
\end{table}

We investigated the impact of the quantity of data on our automatic animal detector by repeating the experiment using four different fractions of the dataset. We randomly subsampled the dataset to $20\%$, $40\%$, $60\%$ and $20\%$ of its original size and in each case we repeated the experiment for 10 different random samples. For all experiments we maintained the $70\%:30\%$ split between training and validation. The results of this experiments, plotted in Figure~\ref{fig:detection}.a), demonstrate that indeed all metrics  improve when more data is used. However, even with only $20\%$ of the dataset, all metrics are above $90\%$. 

To understand the importance of selecting the proportion of images to use in training, we performed two experiments. First, we fixed the number of validation images to $30\%$ (2721 images) and compared the results of training with $20 - 80\%$ of the training set (6349 images). For each measurement, we averaged ten random repetitions, as before. The result (Figure~\ref{fig:detection}.b) shows that although the trend is increasing, the differences diminish faster than when the test and training set were increased in volume. This suggests that perhaps there are difficult cases in the validation set which are less detrimental to average results when the validation set is large. Finally, we tested the performance of our animal detector (Figure~\ref{fig:detection}.c) on various fractions of training and validation sets and found that $70\%:30\%$ is a good choice.

\subsection{Species identification}
We tested two methods for identifying the species of animals in camera trap images. First, as with animal detection, we applied a pre-trained AlexNet in conjunction with an SVM layer that is trained on our dataset. The accuracy for all species except Chital and Muntjac is over $90\%$ using this solution. We experimented with first running an animal detector on the input images, followed by species identification. Indeed, this improved the accuracies for these two classes to $95.47\%$ and $96.19\%$, but accuracies for a few other species (Sambar, Elephant and Wild Pig) dropped significantly.  Our hypothesis is that there is insufficient data to effectively train a classifier for species identification without spatial information about where animals are in the images.

\begin{figure}[htbp]
  \begin{center}
  \begin{tabular}{ l c c c c c c c c c c c | r}
    \rowcolor{up}
        \cellcolor{diag}TP (diagonal)  &\color{black} B &\color{black} C &\color{black}D &\color{black}E &\color{black}G &\color{black}L &\color{black}M &\color{black}S &\color{black}T &\color{black}W &\color{black}U  & \textit{TP+FP} \vspace{.1em}\\
    \cellcolor{left}\color{black}Bear & \cellcolor{diag}108 & 0	 & 1 & 3 & 5 & 1 & 1 & 0 & 2 & 4 & 1 & 126\\
    \cellcolor{left}\color{black}Chital & 0  & \cellcolor{diag}320 & 2 & 1 & 0 & 1 & 4 & 3 & 0 & 1 & 7 & 339\\
    \cellcolor{left}\color{black}Dhole & 2  & 0   & \cellcolor{diag}309 & 2 & 7 & 1 & 7 & 0 & 4 & 0 & 2 & 334\\
    \cellcolor{left}\color{black}Elephant & 6  & 1  & 2  & \cellcolor{diag}302 & 5 & 3 & 0 & 6 & 10 & 4 & 12 & 351\\
    \cellcolor{left}\color{black}Gaur & 6  & 1  & 2  & 6  & \cellcolor{diag}278 & 2 & 2 & 22 & 2 & 4 & 4 & 329\\
    \cellcolor{left}\color{black}Leopard & 1 & 0 & 0 & 1 & 0 & \cellcolor{diag}316 & 0 & 1 & 1 & 0 & 1 & 321\\
    \cellcolor{left}\color{black}Muntjac & 0 & 0 & 2 & 0 & 1 & 1 & \cellcolor{diag}39 & 3 & 0 & 0 & 0 & 46\\
    \cellcolor{left}\color{black}Sambar & 2 & 1 & 4 & 1 & 12 & 0 & 3 & \cellcolor{diag}133 & 1 & 3 & 1 & 161\\
    \cellcolor{left}\color{black}Tiger & 2 & 0 & 1 & 4 & 15 & 3 & 0 & 9 & \cellcolor{diag}305 & 2 & 3 & 344\\
    \cellcolor{left}\color{black}Wild pig & 4 & 0 & 4 & 2 & 4 & 0 & 0 & 2 & 2 & \cellcolor{diag}133 & 5 & 156\\
    \cellcolor{left}\color{black}Unclassified & 0 & 5 & 1 & 6 & 1 & 0 & 4 & 2 & 1 & 4 & \cellcolor{diag}292 & 316 \vspace{.1em}\\ 
    \hline     
    \cellcolor{left}\textit{TP+FN} & 131 & 328 & 328 & 328 & 328 & 328 & 60 & 181 & 328 & 155 & 328 & \\
    \end{tabular}
    \caption{\label{fig:identification} Confusion matrix resulting from AlexNet for species identification. }
  \end{center}
\end{figure}

As a second method, we adapted a recent method developed for object detection given image-level labels~\cite{bilen}. This method, called Weakly Supervised Deep Detection Network (WSDDN), introduces a spatial pyramid pooling layer on top of AlexNet's convolutional layers. The output of these layers is then used in parallel to perform \textit{recognition} over multiple rectangular regions in the image and \textit{detection} of the rectangular region of the image that contains most of the salient information assocated with that image-level label associated with the image. 

Originally, image-level classification scores are obtained by summing the region scores over all regions in each image. This approach of Bilen and Veldadi tends to consider only one (or few) strongly predicted 'tiger' region  equally with several regions that weakly predicted the presence of a 'tiger'. We replace this with the following approach.   We identify the maximum score amongst all classes for each rectangular region and select the top 30 regions based on this maximum class score. Then, we average the score for each class over these 30 regions and pick the class with the highest mean (top-1 class) as the predicted species. Table~\ref{table:AlexWSDDN} shows the performances obtained in spacies classification with Alexnet pre-trained model and with WSDDN model. 

\begin{table*}[ht]
\centering
\begin{tabular}{| l | c c c c|}
 \hline
 \rowcolor{up}
\color{black}Species & \color{black}BC+AlexNet & \color{black}AlexNet & \color{black}WSDDN top-1 & \color{black}WSDDN top-5\\
 \hline
 \cellcolor{left}\color{black}Bear(B) & $89.97\%$ & $93.90\%$ & $98.55\%$ & $99.85\%$\\
 \cellcolor{left}\color{black}Chital(C) & $95.47\%$ & $83.61\%$ & $98.90\%$ & $99.89\%$\\
 \cellcolor{left}\color{black}Dhole(D) & $95.85\%$ & $93.71\%$ & $98.51\%$ & $99.85\%$\\
 \cellcolor{left}\color{black}Elephant(E) & $84.55\%$ & $96.43\%$ & $98.25\%$ & $99.89\%$\\
 \cellcolor{left}\color{black}Gaur(G) & $91.51\%$ & $96.34\%$ & $97.39\%$ & $99.85\%$\\
 \cellcolor{left}\color{black}Leopard(L) & $95.29\%$ & $96.89\%$ & $98.77\%$ & $99.78\%$\\
 \cellcolor{left}\color{black}Muntjac(M) & $96.19\%$ & $74.23\%$ & $98.45\%$ & $99.78\%$\\
 \cellcolor{left}\color{black}Sambar(S) & $88.62\%$ & $95.62\%$ & $97.48\%$ & $99.89\%$\\
 \cellcolor{left}\color{black}Tiger(T) & $96.14\%$ & $90.59\%$ & $98.36\%$ & $99.89\%$\\
 \cellcolor{left}\color{black}Wild Pig(W) & $76.11\%$ & $93.06\%$ & $97.91\%$ & $99.96\%$\\
 \cellcolor{left}\color{black}UnClassified(U) & $-$ & $98.94\%$ & $98.41\%$ & $99.85\%$\\ 
 \hline
\end{tabular}
\caption{A comparison of Binary Classification (BC) combined with AlexNet and WSDDN using top-$1$ and top-$5$ classes.}
\label{table:AlexWSDDN}
\end{table*}

\begin{figure}[htbp]
  \begin{center}
   \begin{tabular}{@{}c@{}c@{}c@{}c@{}c@{}}
      \includegraphics[width=.19\linewidth]{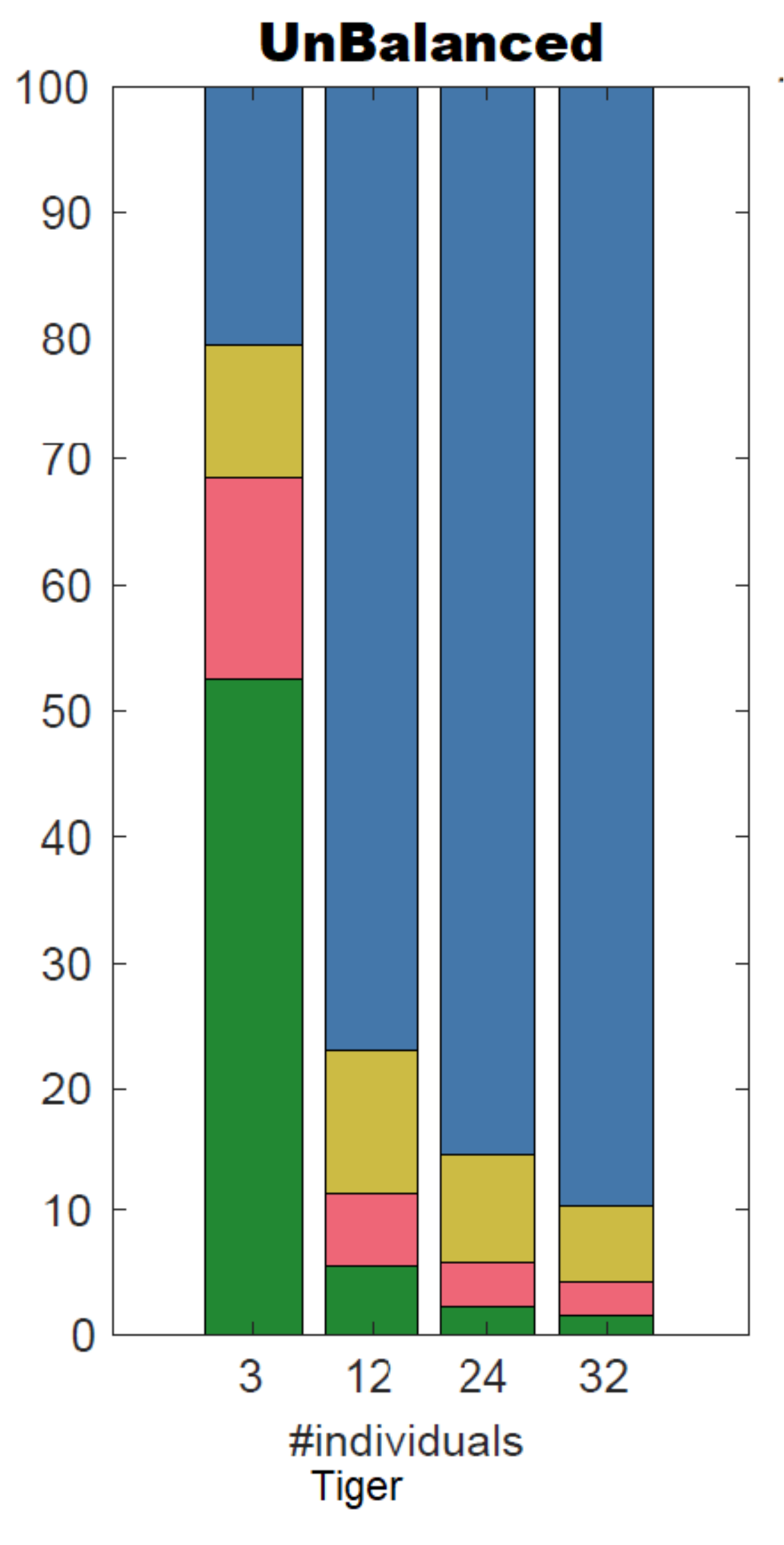} &
      \includegraphics[width=.19\linewidth]{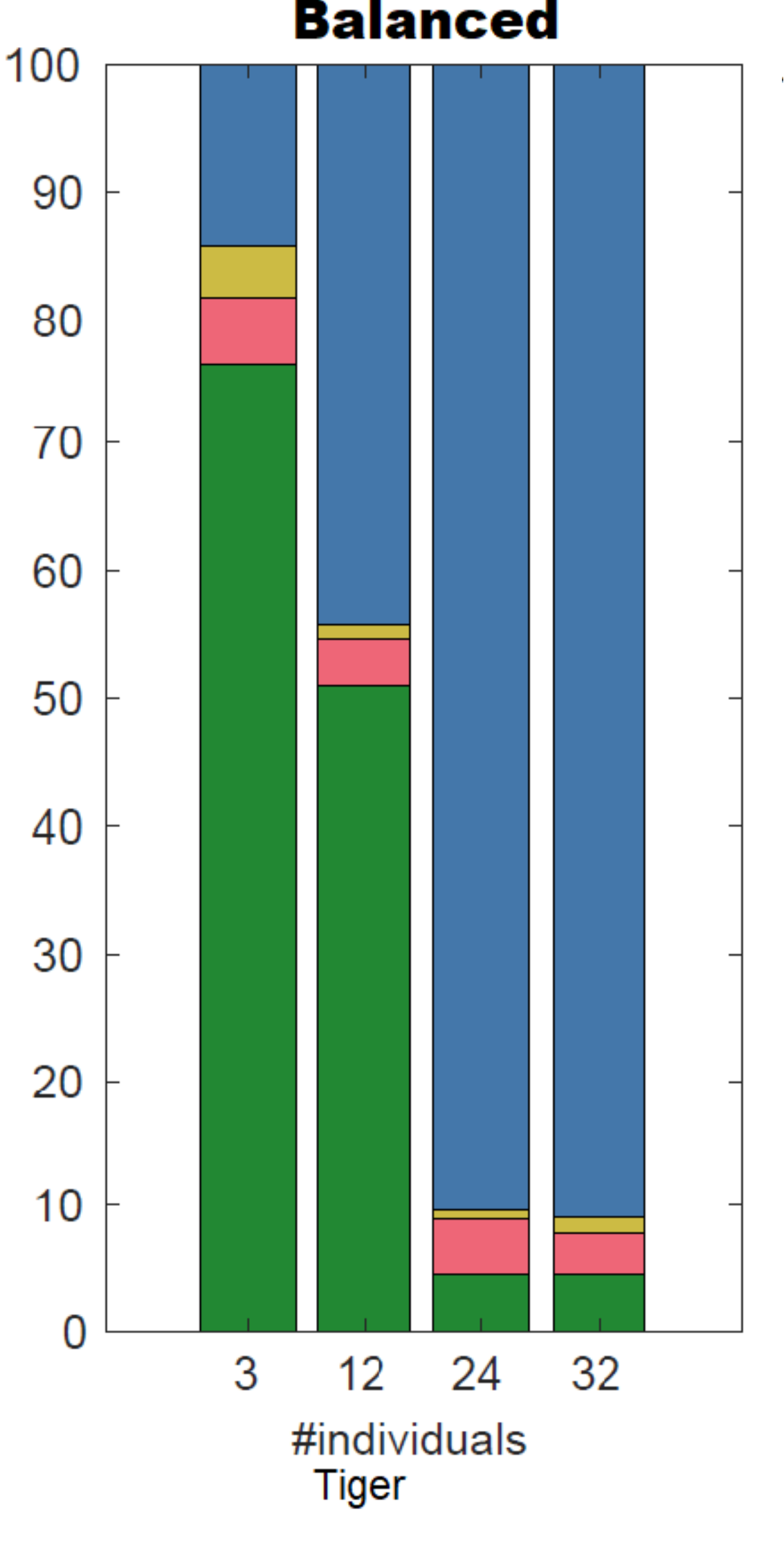} &
      \includegraphics[width=.19\linewidth]{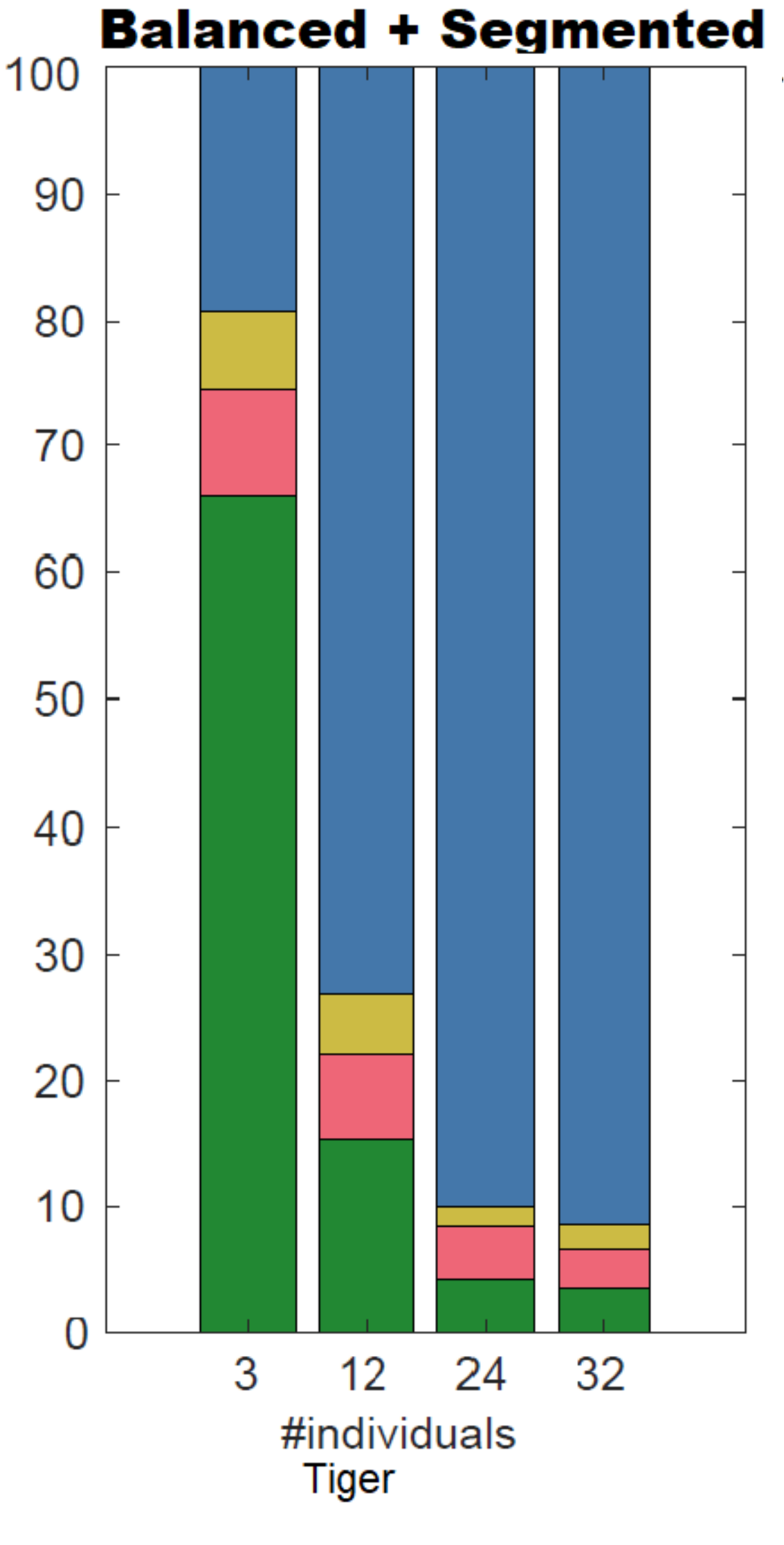} & 
      \includegraphics[width=.19\linewidth]{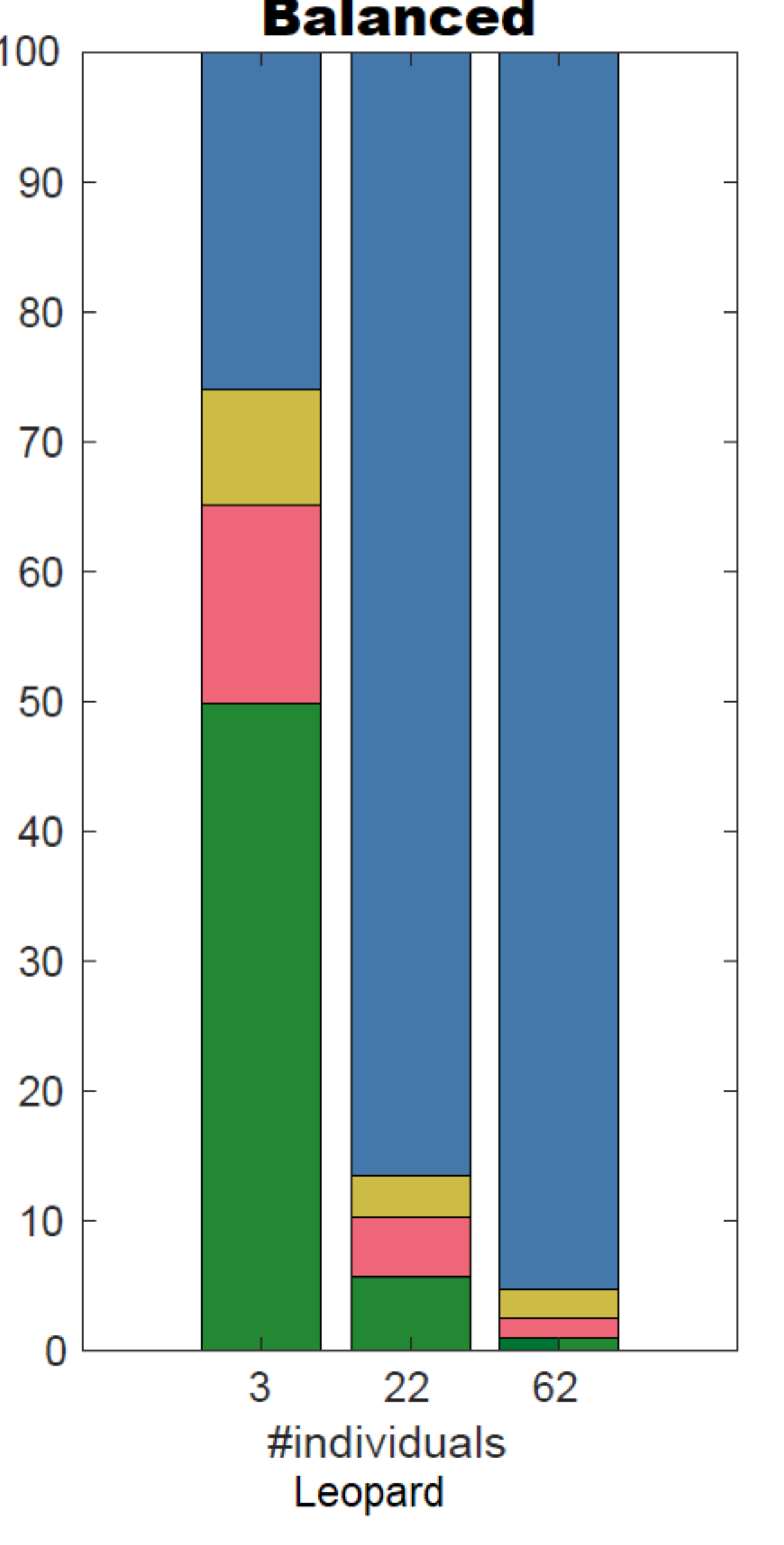} &
      \includegraphics[width=.19\linewidth]{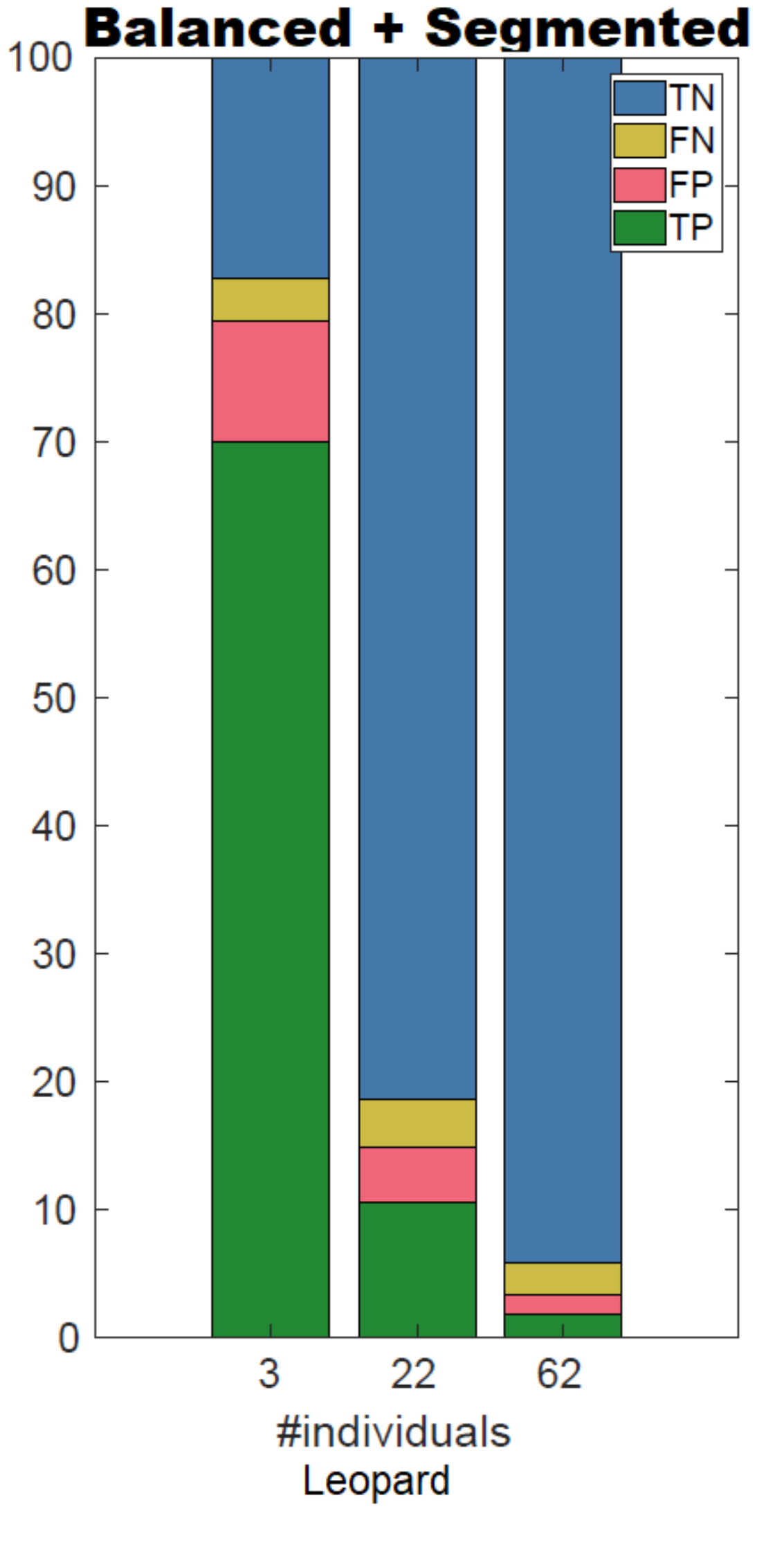} \\
      \multicolumn{3}{c}{a) recognising individual tigers}&
      \multicolumn{2}{c}{b) recognising individual leopards}
    \end{tabular}
    \caption{\label{fig:recognition}Individual recognition results. }
  \end{center}
\end{figure}

\subsection{Individual recognition of tiger and leopard}

We applied the same architecture (WSDDN) to recognise individual animals by treating each individual (rather than species) as a separate class. That is, the pre-trained network was fine-tuned using several different images of known individuals to learn a mapping from characteristic markings on the skin to the specific individuals with those markings. For this, we used the labelled individuals in the dataset. We tested the efficacy of individual recognition using by training with different combinations: Only tigers, only leopards, and leopards and tigers combined. 

For the case where we trained using only one class (leopards or tigers), we investigated the impact of balancing the dataset so that each class (individual) was represented equally during training. In addition, motivated by the importance that segmentation plays in vision-based tasks~\cite{shukla}, we explored the use of segmented animals to train our networks. We segmented images automatically using a unary classifier (our animal detector on patches) whose result is used to obtain a segmentation mask by considering pair-wise similarities between patches~\cite{krahenbuhl}. The numbers of true negatives, false negatives, false positives and true positives for the various combinations are shown in figure~\ref{fig:recognition}.
\begin{wrapfigure}{R}{.5\linewidth}
        \includegraphics[width=\linewidth]{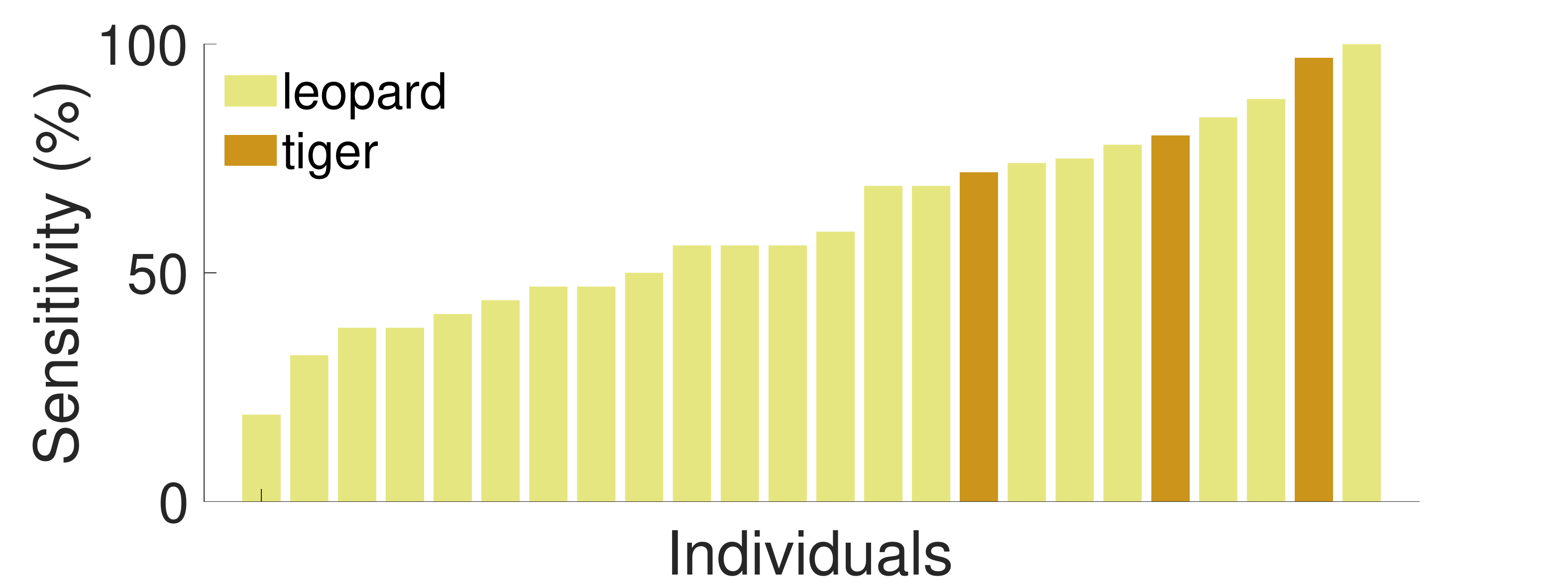}
        \caption{Sensitivities of individual recognition for $24$ leopard and tiger individuals after being trained jointly.}
    \label{fig:jointindividual}
\end{wrapfigure}

To create a balanced dataset across all individuals, dozens of images were removed so that every individual was represented using the same number of images in the training data. Despite this reduction of total data, we found that balancing the training dataset significantly improves all metrics for tigers as well as leopards (Figure~\ref{fig:recognition}) . More surprisingly, we observed that training using automatically segmented tigers did not improve the classifier.

We tested a jointly trained classifier that is able to recognize individuals of leopard and tiger. We chose the $3$ tiger individuals and $21$ leopard individuals for which we had more than $100$ images each and trained a classifier to recognize any of the $24$ individuals. We found the accuracy and specificity of all individuals to be high ($>95\%$). However, the sensitivity of the classifier varied widely across the individuals (see figure~\ref{fig:jointindividual}). We attribute the differences to the varying qualities of the images for different individuals.

\bibliography{nee18-rita}










\end{document}